\pdfoutput=1

\documentclass[11pt]{article}

\usepackage[]{ACL2023}

\usepackage{times}
\usepackage{latexsym}

\usepackage[T1]{fontenc}

\usepackage[utf8]{inputenc}

\usepackage{microtype}

\usepackage{inconsolata}

\usepackage{graphicx}
\usepackage{amsmath}
\usepackage{multirow}
\usepackage{hhline}
\usepackage{pifont}
\usepackage{enumitem}

\usepackage{hyperref}

\usepackage[ruled, lined, commentsnumbered, longend]{algorithm2e}
\usepackage{xcolor}

\newcommand\mycommfont[1]{\small\ttfamily\textcolor{teal}{#1}}
\SetCommentSty{mycommfont}

\usepackage{etoolbox}
\usepackage{color}
\makeatletter
\providecommand{\subtitle}[1]{
  \apptocmd{\@title}{
    \par\medskip 
    \small\textcolor[rgb]{0.66,0,0}{\textit{#1}}\par 
  }{}{}
}
\makeatother

\newcommand\blfootnote[1]{%
  \begingroup
  \renewcommand\thefootnote{}\footnote{#1}%
  \addtocounter{footnote}{-1}%
  \endgroup
}

\title{Label-aware Hard Negative Sampling Strategies with Momentum Contrastive Learning for Implicit Hate Speech Detection}
\subtitle{Warning: This paper contains examples that can be offensive or upsetting.}

\author{Jaehoon Kim\textsuperscript{1}, Seungwan Jin\textsuperscript{2}, Sohyun Park\textsuperscript{1}, Someen Park\textsuperscript{1}, Kyungsik Han\textsuperscript{1,2,*}\\ 
\textsuperscript{1} Department of Artificial Intelligence, Hanyang University, Seoul, Republic of Korea \\ 
\textsuperscript{2} Department of Data Science, Hanyang University, Seoul, Republic of Korea \\
\texttt{\{jaehoonkimm,seungwanjin,sohyunpark,someeeen,kyungsikhan\}@hanyang.ac.kr} \\
}

\begin{document}
\maketitle

\blfootnote{*Corresponding author}

\begin{abstract}
    Detecting implicit hate speech that is not directly hateful remains a challenge. Recent research has attempted to detect implicit hate speech by applying contrastive learning to pre-trained language models such as BERT and RoBERTa, but the proposed models still do not have a significant advantage over cross-entropy loss-based learning. We found that contrastive learning based on randomly sampled batch data does not encourage the model to learn hard negative samples. In this work, we propose \textbf{La}bel-aware \textbf{H}ard \textbf{N}egative sampling strategies (\textbf{LAHN}) that encourage the model to learn detailed features from hard negative samples, instead of naive negative samples in random batch, using momentum-integrated contrastive learning. LAHN outperforms the existing models for implicit hate speech detection both in- and cross-datasets. The code is available at \href{https://github.com/Hanyang-HCC-Lab/LAHN}{https://github.com/Hanyang-HCC-Lab/LAHN}
\end{abstract}
\section{Introduction}
Online hate speech has become one of the major social problems as it leads to discrimination against certain groups and social conflict~\cite{howard2019free,matamoros2021racism}. Hate speech can be either explicit, which directly uses hateful language, or implicit, which metaphorically implies hateful language~\cite{waseem2017understanding}. While explicit hate speech can be addressed relatively easily through the use of automated filters~\cite{watanabe2018hate,xiang2012detecting,rodriguez2020automatic}, 
dealing with implicit hate speech is highly challenging. For example, identity term bias~\cite{elsherief2021latent,dixon2018measuring}, where identity terms (e.g., black, jew) frequently appear in hateful contexts, can over-bias the model and cause false positives in implicit hate speech detection. Given the characteristics of implicit hate speech, it will be important to encourage models to learn subtle differences between similar sentences that might otherwise confuse them. Research has built implicit hate speech datasets of implicit hate speech~\cite{elsherief2021latent,sap2019social,hartvigsen2022toxigen,vidgen2020learning} and proposed detection models~\cite{kim2022generalizable,kim2023conprompt} using contrastive learning; however, the model showed limited performance improvement in in-dataset evaluation or have limitations that require external knowledge or additional computational resources for pre-training.

\begin{figure}
\centering
\includegraphics[width=1.0\columnwidth]{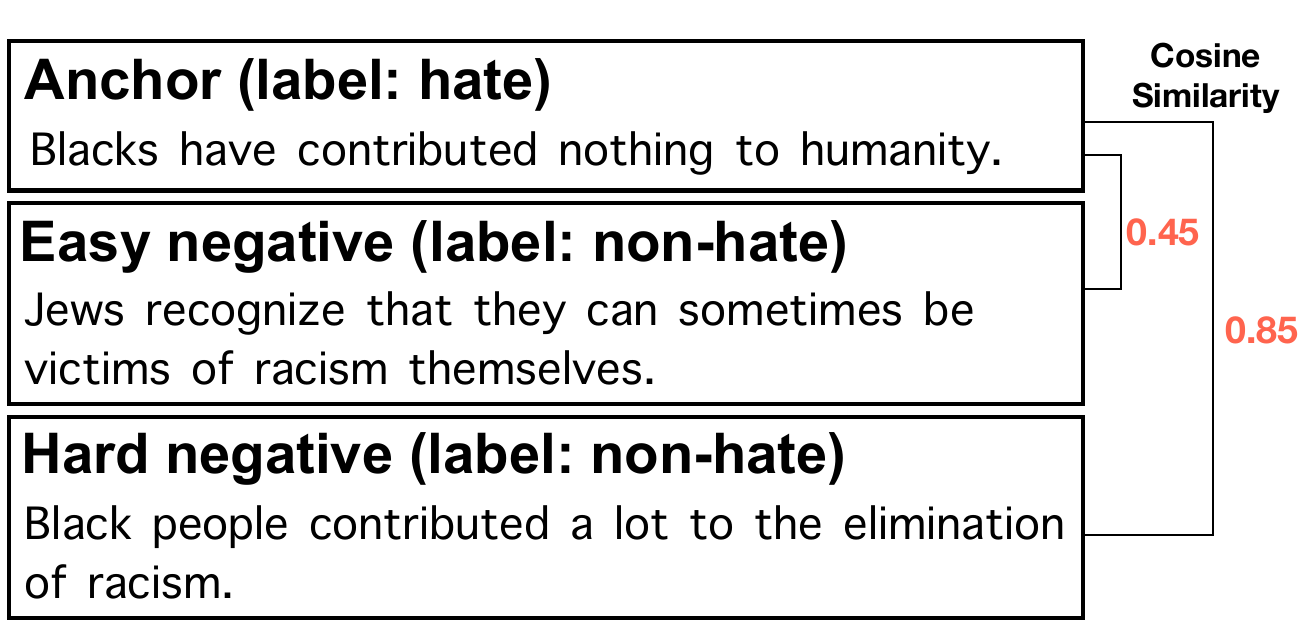}
\caption{Our research motivation. Easy negatives have a low semantic similarity to anchors, and hard negatives have a high semantic similarity to anchors.}
~\label{fig1}
\vspace{-0.85cm}
\end{figure}

Figure~\ref{fig1} shows an easy negative (middle) and a hard negative (bottom) for an anchor sentence targeting black. Easy negatives have opposite labels and are relatively distinct from the anchor, meaning that for the model it is not difficult to distinguish from the anchor. On the other hand, hard negatives are semantically similar to the anchor but have opposite labels, meaning that the model may have difficulty distinguishing them from the anchor. The key task is to effectively identify hard negatives and train the model accordingly. If the negative sample trained along with the anchor is similar to the anchor, the model can be trained to better distinguish relatively indistinguishable data~\cite{robinson2020contrastive,jiang2022supervised,wu2020conditional}. However, existing work has used naive contrastive learning, which encourages contrast between data in randomly sampled mini-batches to learn representations and fails to guarantee the learning of hard negative samples.

\begin{figure}
\centering
\includegraphics[width=0.95\columnwidth]{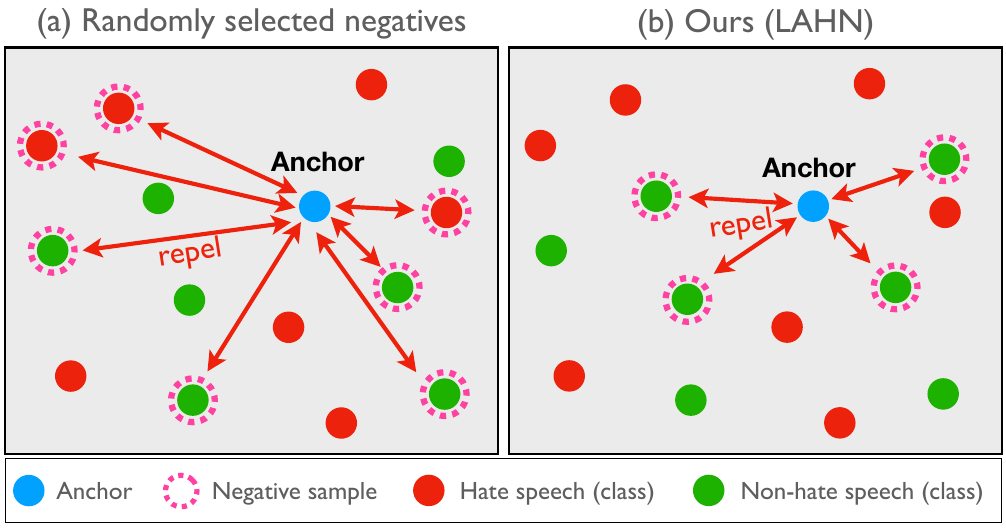}
\caption{Illustration of the two methods using contrastive learning in a situation where the class of the anchor is hate speech. (a) The random sampling method randomly selects negative samples. (b) LAHN selects only different classes from the anchor as negative samples (i.e., the green dots include only the opposite class from the anchor).}
~\label{fig2}
\vspace{-0.6cm}
\end{figure}
In this work, we propose a novel approach to implicit hate speech detection, namely \textbf{La}bel-aware \textbf{H}ard \textbf{N}egative sampling strategies (\textbf{LAHN}). LAHN focuses more on distinguishing between the anchor and hard negatives, mitigating overfitting to the context of the text or specific words. Inspired by MoCo~\cite{he2020momentum}, LAHN employs a momentum queue to effectively expand the negative samples that are candidates for hard negative samples, and compactly performs contrastive learning by extracting the top-$k$ hard negatives that require sophisticated disentangling from the anchor. A key difference from previous research is that LAHN extracts hard negatives from the momentum queue in contrastive learning based on the similarity between the anchor and the negatives and the level of ambiguity for each negative (Figure~\ref{fig2}).

In summary, our contributions are as follows:
\begin{itemize}
\setlength{\itemsep}{0pt}
    \item We propose LAHN, a novel method that focuses on hard negatives that should be disentangled from the anchor to promote the effective learning of hate speech representations with implicit characteristics.
    \item Contrary to previous studies, we observe that LAHN can improve performance in both in- and cross-dataset evaluation with a simple dropout noise augmentation without external knowledge and additional cost.
    \item We validate the generalized learning effect of LAHN by achieving state-of-the-art performance in both in- and cross-dataset evaluation on four representative public benchmark datasets for implicit hate speech detection.
\end{itemize}

\begin{figure*}
\centering
\includegraphics[width=2.0\columnwidth]{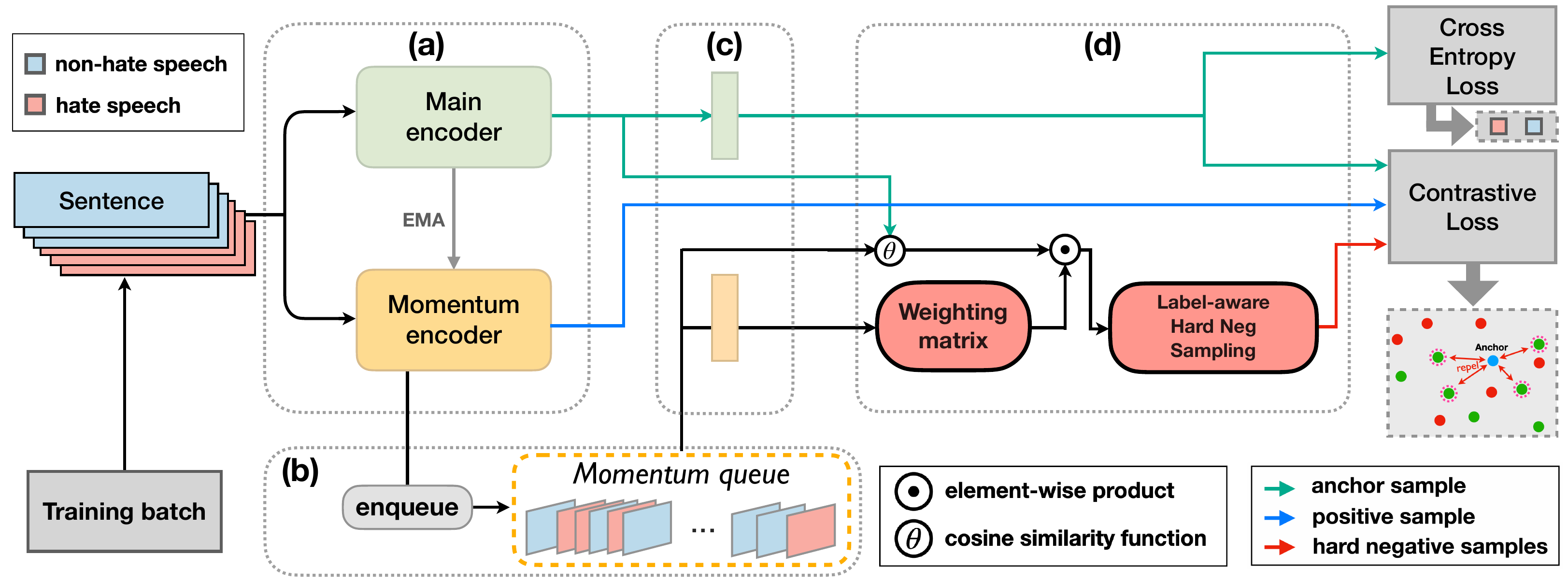}
\caption{The overview of our LAHN. (a) shows the Momentum encoder being updated via EMA based on the main encoder. (b) enqueues the features extracted by the Momentum encoder. (c) is the prediction head of each encoder, which returns the prediction logits of the input features. (d) weights the true negatives in the momentum queue based on the prediction probability obtained from (c) and samples the hard negatives among them. ($\odot$: element-wise product function, $\theta$: cosine similarity function.)}
~\label{fig3}
\vspace{-0.5cm}
\end{figure*}

\section{Related work}
\subsection{Implicit Hate Speech Detection}
Implicit hate speech refers to expressions of hatred or discrimination that are not directly stated or overtly aggressive but are conveyed through subtle, indirect language, insinuations, or coded messages~\cite{elsherief2021latent,kim2023conprompt}. According to prior studies~\cite{ocampo2023depth,yang2023hare}, it is challenging to identify implicit hate speech due to its reliance on background knowledge, cultural context, and the ability to infer the implied meanings behind seemingly neutral or ambiguous expressions. For instance, ~\citet{ocampo2023depth} explores the absence of consistent hate-speech-specific prosody across languages, indicating the importance of linguistic and cultural nuances in understanding hate speech.

For this reason, the existing pre-trained language models in the hate speech domain, such as HateBERT~\cite{caselli2021hatebert} or fBERT~\cite{sarkar2021fbert}, may have a spurious correlation issue that classifies an input text as hateful due to the presence of specific identity terms (e.g., Black, Asian, etc.) and this hinders generalized performance. This body of work underscores the need for sophisticated, context-aware approaches in hate speech detection systems to effectively address the subtleties of implicit hate speech, considering its varied manifestations and significant societal impact.

Contrastive learning methods~\cite{wu2020clear,he2020momentum,khosla2020supervised,giorgi2020declutr,gao2021simcse} has been increasingly recognized as a necessary and effective approach for detecting implicit hate speech due to its ability to handle the context-dependent nature of such content by leveraging the subtle differences and similarities between hateful and non-hateful content, enhancing model sensitivity to the nuanced expressions of hate speech. ~\citet{kim2022generalizable} proposed ImpCon, which uses the external knowledge (i.e., the implication of anchor sentences) as positive samples using contrastive loss for implicit hate speech detection. The recently proposed ConPrompt~\cite{kim2023conprompt} utilized machine-generated statements to improve implicit hate speech performance by applying contrastive learning using example sentences from the original prompt as positive samples. However, these previous methods still have limitations due to reliance on pre-defined external knowledge or text generation costs.

Recently, there has been research in using Large Language Models (LLMs) for hate speech detection.~\citet{zhang2024don} investigated the sensitivity and limitations of LLMs for performing hate speech detection, and ~\citet{roy2023probing} explored several prompting strategies to improve the detection capabilities of LLMs.~\citet{yang2023hare} proposed the HARE framework, which incorporates explanations generated using chain-of-thoughts (CoT) into the language model learning process. Despite these efforts, the performance of LLMs in hate speech detection is still limited, and costly compared to the language models such as BERT and RoBERTa. Our experimental results on implicit hate speech detection using LLMs are presented in Appendix~\ref{sec:appendix}.

\subsection{Hard Negative Mining}
Hard negative mining is an important technique in various areas of machine learning that significantly improves a model's performance by carefully selecting negative samples that are less obviously distinguishable from positive samples~\cite{robinson2020contrastive,gunel2020supervised,schroff2015facenet,wu2020conditional,ge2021robust,kalantidis2020hard}. In contrastive learning, many studies have found that focusing on hard negative samples among negative samples learns better representations and improves performance~\cite{robinson2020contrastive,gunel2020supervised,schroff2015facenet,kalantidis2020hard}. 

In the field of computer vision, many studies related to hard negatives have been proposed. ~\citet{ge2021robust} identified the impact of hard negatives in learning strategies to generate appropriate negative samples, and ~\citet{robinson2020contrastive} proposed H-SCL, which uses the label information of samples to introduce hard negative sampling in supervised contrastive learning. ~\citet{wu2020conditional} proposed a ring method to step-wise retrieve hard negative samples that are neither too hard nor too easy, and experiments showed that dynamically changing the hard negative sampling range can help to learn.

In NLP, hard negative mining can help models learn finer distinctions between textual data and better understand the nuances of language, context, and semantic meaning. The supervised SimCSE for learning text representations proposed by ~\citet{gao2021simcse} considered predefined contradiction pairs as hard negatives for the first time in the field of NLP and encouraged learning to distinguish hard negatives in contrastive learning. However, supervised SimCSE relies on predefined in-batch hard negatives and only utilizes samples within the batch in contrastive learning, so it is limited in considering negative samples broadly. ESimCSE~\cite{wu2022esimcse} and MoCoSE~
\cite{cao2022exploring}, which were proposed to compensate for these limitations of SimCSE, integrated momentum contrastive learning with SimCSE, but hard negative samples were not considered in the contrastive learning process.

To the best of our knowledge, no research has explored how to effectively integrate hard negatives with supervised contrastive learning using label information without relying on external knowledge. In this paper, we propose LAHN, which integrates momentum contrastive learning and label-aware hard negative sampling strategies to effectively handle implicit hate speech data.
\vspace{0.25cm}

\setlength\floatsep{1.25\baselineskip plus 3pt minus 2pt}
\setlength\textfloatsep{1.0\baselineskip plus 3pt minus 2pt}
\setlength\intextsep{1.25\baselineskip plus 3pt minus 2 pt}

\vspace{-0.3cm}

\begin{algorithm}[t]
    \mycommfont{\# E, E\_m: main encoder end momentum encoder} \\ 
    \mycommfont{\# queue: momentum contrast queue} \\
    \mycommfont{\# m, t: momentum and temperature parameter} \\
    \mycommfont{\# k: hard negative sampling size} \\
    \mycommfont{\# sim: cosine similarity function} \\
    
    \mbox{}\\
    \mycommfont{\# load a batch x with N samples} \\
    for x in loader:  \\ 
    \quad x\_anc, pred = E.forward(x) \\
    \quad x\_aug, \_ = E\_m.forward(x) \\ 
    \quad x\_aug = x\_aug.detach() \mycommfont{\# no gradient} \\ 
    \mbox{}\\
    \quad \mycommfont{\# enqueue the current batch embeddings} \\
    \quad enqueue(queue, x\_aug)  \\
    \quad \mycommfont{\# dequeue the earliest batch embeddings} \\
    \quad dequeue(queue) \\ 
    \mbox{}\\
    \quad \mycommfont{\# hard negative sampling for anchor} \\
    \quad \_, weights = E\_m.forward(queue) \\
    \quad negs = sort(sim(x\_anc, queue) * weights) \\
    \quad hard\_neg = topk(negs, k) \mycommfont{\# top-k sampling} \\
    \quad \mycommfont{\# extract pos and neg simialrity} \\
    \quad pos = sim(x\_anc, x\_aug)/t \\ 
    \quad pos = diag(pos).view(-1, 1) \mycommfont{\# extract pos} \\ 
    \quad neg = sim(x\_anc, hard\_neg)/t \\
    \quad logits = concat([pos, neg], dim=1) \\
    \mbox{}\\
    \quad \mycommfont{\# Contrastive loss and CE loss} \\ 
    \quad labels = Zeros(logits)[:, 0] = 1 \\
    \quad loss\_1 = CrossEntropy(logits, labels) \\ 
    \quad loss\_2 = CrossEntropy(pred, x.labels) \\
    \quad (loss\_1 + loss\_2).backward() \\ 
    \quad update(E.params)
    \mbox{}\\
    \quad\mycommfont{\# momentum update} \\
    \quad E\_m.params = m*E\_m.params+(1-m)*E.params \\ 
    
    \caption{\textbf{Pseudocode of LAHN}}
    \label{algo:1}
\end{algorithm}

\section{Method}

As shown in Figure~\ref{fig2}-(a), the existing method performs contrastive learning on all other samples with equal weight, including those of the same class, for a given anchor in a mini-batch. This characteristic makes it difficult for the model to focus on distinguishing between the anchor and hard negatives, which are semantically similar and embedded close to the anchor.

On the other hand, our LAHN first uses label information to identify negatives with labels different from the anchor's, which should be embedded in opposite directions. Next, the momentum encoder calculates the probability that the negative is in the same class as the anchor, and this probability is used to multiply the similarity value between the negative and the anchor. Finally, LAHN extracts the top-k hard negatives based on the calculated values and performs compact contrastive learning for the given anchor (Figure~\ref{fig2}-(b)). Through this, LAHN better captures the detailed differences between implicit hate speech and non-hate speech by focusing on hard negatives.

Figure~\ref{fig3} illustrates the overview of our LAHN. Random mini-batch samples are used as the input to the main and momentum encoders, and the output embedding of the momentum encoder is inserted into the momentum queue. Hard negative sampling for each anchor is performed after the momentum queue is filled with at least a quarter of the size. As in MoCo~\cite{he2020momentum}, we use an exponential moving average (EMA) to slowly update the momentum encoder based on the weights of the main encoder. Algorithm~\ref{algo:1} presents the pseudocode implementation.
 
\subsection{Label-aware Hard Negative Sampling} \label{sec:Main_Approach}
In this section, we propose a label-aware hard negative sampling strategy to effectively integrate label information and hard negative mining methods in supervised contrastive learning. We assume that focusing on contrastive learning to distinguish the anchor from the hard negatives that are semantically similar to the anchor can help learn disentangled representations of implicit hate speech data.

In supervised contrastive learning, false negatives, where samples with the same label as the anchor are used as negative samples, can hinder the representation learning of the model~\cite{kalantidis2020hard}. To exclude false negative cases in the hard negative sampling process, we identified the true negatives of the momentum queue using the label information of each anchor in the batch and then sorted them by finding the cosine similarity between anchors and true negatives.

To sample hard negatives, which should be primarily included for successful compact contrastive learning, we considered not only the contextual similarity to the anchor, but also the ambiguity of the representation to classify the class of the hard negative. First, we calculated the probability that all true negatives in the momentum queue, which are hard negative candidates for contrastive learning, are predicted as the class of the anchor through the prediction head of the momentum encoder. Second, we applied hard negative sampling based on the assumption that the closer the probability of being predicted as the anchor class for true negatives is to 1 (more ambiguity), the more sophisticated representation learning is required. Finally, we multiplied the prediction probabilities by the similarity values with the anchor, sampled the top-k hard negatives based on the multiplied values, and assigned them as negatives of each anchor in compact contrastive learning.

\subsection{Training objective for Implicit Hate Speech Detection}
In this section, we conduct contrastive learning for disentangled representation learning and classification training for implicit hate speech detection. First, we employ momentum contrastive learning~\cite{he2020momentum} to maximize the effectiveness of our proposed hard negative sampling strategy. In the implicit hate speech domain, text augmentations, such as the most commonly used token replacement (e.g., synonym substitution), can alter a sentence's hateful or non-hateful nature. Therefore, we utilized the augmentation technique that employs dropout noise, as used in SimCSE~\cite{gao2021simcse}. 
In Section~\ref{results}, the experiments using external knowledge employed the augmentation techniques (i.e., implication on hate speech and synonym substitution on non-hate speech) used in~\cite{kim2022generalizable}. After hard negative sampling in Section~\ref{sec:Main_Approach}, we used InfoNCE~\cite{oord2018representation} loss for the contrastive learning as follows:
\begin{align}
\mathcal{L}_{\text{CL}} = -\log \frac{\exp(\text{sim}(x_i, x_i^p))/\tau}{\sum_{j=1}^N \exp(\text{sim}(x_i, x_j^n))/\tau}
\end{align}
where $x_i$ is the anchor sample of the batch, and every anchor has one positive sample $x^p$. The proposed method uses only the batch's positive samples and the hard negatives sampled from the momentum queue as negatives. $N$ denotes the number of selected hard negatives from the queue, $x^n$ denotes the negative samples, $\tau$ is a temperature parameter, and $sim$ is the cosine similarity function. To minimize the loss in the corresponding contrastive learning process, the similarity between anchor and positive samples should be maximized, and the similarity between anchor and negative samples should be minimized.
The loss function for implicit hate speech detection is as follows.
\begin{align}
\mathcal{L}_{\text{CE}} = -\frac{1}{N}\sum^{N}_{i=1}{[y_{i}\,\log{\hat{y}}_{i} + (1-y_{i}) \log{(1 - \hat{y}_{i})}]}
\end{align}
where $\hat{y}_{i}\in\{0,1\}$ is one-hot encoded ground truth. The final loss that we use for training is as follows:
\begin{align}
    L = (1-\lambda) L_{CL} + \lambda L_{CE}
\end{align}
where $\lambda$ is a loss scaling parameter and we empirically found and set the parameter $\lambda$ as 0.1.
\newcommand{\cmark}{\ding{51}}%
\newcommand{\xmark}{\ding{55}}%
\newcommand{\bluecheck}{{\color{blue}\cmark}}
\newcommand{\redxmark}{{\color{red}\xmark}}

\begin{table*}[]
\centering
\caption{To investigate both case with/without external knowledge, we employ two datasets (IHC, SBIC) with implication information as the training dataset among the public benchmark dataset. The top table shows the results of the in-dataset (IHC) evaluation and the cross-dataset (SBIC, DynaHate, ToxiGen) evaluation. The bottom table contains the results of the in-dataset (SBIC) evaluation and the cross-dataset (IHC, DynaHate, ToxiGen) evaluation.}
\label{table_1}
\resizebox{0.95\textwidth}{!}{
\begin{tabular}{c|c|c|l|cccc|c}
\hline
\multirow{2}{*}{\begin{tabular}[c]{@{}c@{}}\textbf{Training} \\ \textbf{Dataset}\end{tabular}} &
  \multirow{2}{*}{\begin{tabular}[c]{@{}c@{}}\textbf{Pre-trained} \\ \textbf{Language Model}\end{tabular}} &
  \multirow{2}{*}{\begin{tabular}[c]{@{}c@{}}\textbf{External} \\ \textbf{Knowledge}\end{tabular}} &
  \multicolumn{1}{c|}{\multirow{2}{*}{\textbf{Objective}}} &
  \multicolumn{4}{c|}{\textbf{Evaluation Dataset}} &
  \multirow{2}{*}{\textbf{Average}} \\ \cline{5-8}
 &
   &
   &
  \multicolumn{1}{c|}{} &
  \multicolumn{1}{l|}{\textbf{IHC}} &
  \multicolumn{1}{l|}{\textbf{SBIC}} &
  \multicolumn{1}{l|}{\textbf{DynaHate}} &
  \multicolumn{1}{l|}{\textbf{ToxiGen}} &
   \\ \hline
\multirow{12}{*}{IHC} & \multirow{7}{*}{BERT-base}    & \xmark & Cross-Entropy Loss (CE)           & 77.49          & 57.05          & 53.69     &  60.80   & 62.26          \\
                      &                               & \xmark & SCL~\cite{gunel2020supervised}  & 77.81          & 59.19          & 55.84     &  62.19   & 63.76          \\
                      &                               & \xmark & \textbf{LAHN (ours)}& \textbf{78.40}  & \textbf{62.83} & \textbf{57.80}  &  \textbf{63.21}   &\textbf{65.56}          \\ \cline{3-9} 
                      &                               & \cmark & ImpCon~\cite{kim2022generalizable}    &  78.39   & 54.55  & \textbf{59.41}  &   59.64    & 63.00            \\
                      &       & \cmark & \textbf{LAHN (ours)}& \textbf{78.62}    & \textbf{62.02}    & 56.13    &   \textbf{62.92}   & \textbf{64.92}   \\ \cline{2-9} \hhline{~-=======}
                      & \multirow{7}{*}{RoBERTa-base} & \xmark & Cross-Entropy Loss (CE)           & 79.95          & 55.03          & 47.34     & 59.35     & 60.42\         \\
                      &                               & \xmark & SCL~\cite{gunel2020supervised}  & 79.33          & 57.77       & 46.92   & 60.89    & 61.23          \\
                      &                               & \xmark & \textbf{LAHN (ours)}& \textbf{80.11}  & \textbf{60.57}  & \textbf{48.46}  & \textbf{63.94}  & \textbf{63.27} \\ \cline{3-9} 
                      &                               & \cmark & ImpCon~\cite{kim2022generalizable}    & 78.78     & 63.82          & \textbf{50.13}    & 61.79   & 63.63           \\
                      &                               & \cmark & \textbf{LAHN (ours)}& \textbf{80.58} & \textbf{64.01} & 49.54 & \textbf{64.49} & \textbf{64.66}          \\ \hline
\end{tabular}
}

\vspace{0.05cm}

\resizebox{0.95\textwidth}{!}{
\begin{tabular}{c|c|c|l|cccc|c}
\hline
\multirow{2}{*}{\begin{tabular}[c]{@{}c@{}}\textbf{Training} \\ \textbf{Dataset}\end{tabular}} &
  \multirow{2}{*}{\begin{tabular}[c]{@{}c@{}}\textbf{Pre-trained} \\ \textbf{Language Model}\end{tabular}} &
  \multirow{2}{*}{\begin{tabular}[c]{@{}c@{}}\textbf{External} \\ \textbf{Knowledge}\end{tabular}} &
  \multicolumn{1}{c|}{\multirow{2}{*}{\textbf{Objective}}} &
  \multicolumn{4}{c|}{\textbf{Evaluation Dataset}} &
  \multirow{2}{*}{\textbf{Average}} \\ \cline{5-8}
 &
   &
   &
  \multicolumn{1}{c|}{} &
  \multicolumn{1}{l|}{\textbf{IHC}} &
  \multicolumn{1}{l|}{\textbf{SBIC}} &
  \multicolumn{1}{l|}{\textbf{DynaHate}} &
  \multicolumn{1}{l|}{\textbf{ToxiGen}} &
   \\ \hline
\multirow{12}{*}{SBIC} & \multirow{6}{*}{BERT-base}    & \xmark & Cross-Entropy Loss (CE)        & 59.47    & 83.72     & 60.17    &  67.54  & 67.73          \\
                       &                               & \xmark & SCL~\cite{gunel2020supervised}  & 60.07   & \textbf{84.14}   & 60.97 & 67.62   & 68.20          \\
                       &                               & \xmark & \textbf{LAHN (ours)}& \textbf{62.36}          & 83.98          & \textbf{63.06}     &  \textbf{69.58}   & \textbf{69.75}          \\ \cline{3-9} 
                       &                               & \cmark & ImpCon~\cite{kim2022generalizable} & 58.64         & 83.53          & 59.50      &  66.54  & 67.05   \\
                       &                               & \cmark & \textbf{LAHN (ours)}& \textbf{61.58}          & \textbf{84.31}          & \textbf{60.97}     &  \textbf{68.52}   & \textbf{68.85}          \\ \cline{2-9} \hhline{~-=======}
                       & \multirow{6}{*}{RoBERTa-base} & \xmark & Cross-Entropy Loss (CE)           & 59.68          & 85.27 & 61.62     &  68.54   & 68.78          \\
                       &                               & \xmark & SCL~\cite{gunel2020supervised}  & 59.61          & 85.25 & 61.17     &  68.77   & 68.70          \\
                       &                               & \xmark & \textbf{LAHN (ours)}& \textbf{64.74} & \textbf{85.45} & \textbf{64.32} &  \textbf{70.61}   & \textbf{71.28} \\ \cline{3-9} 
                       &                               & \cmark & ImpCon~\cite{kim2022generalizable} & 56.95 & 84.66 & 60.70     & 66.77    &  67.27         \\
                       &                               & \cmark & \textbf{LAHN (ours)}& \textbf{64.05} & \textbf{85.80} & \textbf{63.26}    &   \textbf{69.91}   & \textbf{70.76} \\ \hline
\end{tabular}
}

\end{table*}

\section{Experiment}
\subsection{Datasets}
Similar to the approach in ~\cite{kim2022generalizable, hartvigsen2022toxigen}, we used the implicit hate speech benchmark datasets---IHC, SBIC, DynaHate, and ToxiGen---for model evaluation. 
\begin{itemize}[leftmargin=0.35cm]
\setlength{\itemsep}{0pt}
    \item \textbf{Implicit Hate Speech Corpus}~\cite{elsherief2021latent}: The dataset focuses on implicit hate speech collected from Twitter. It comprises 19,112 tweets, with 4,909 labeled as implicit hate and 933 labeled as explicit hate.
    \item \textbf{Social Bias Inference Corpus}~\cite{sap2019social}: The dataset with hierarchical categories of social biases and stereotypes. It supports large-scale modeling and evaluation with 150k structured annotations of social media posts, covering over 34k implications about demographic groups. 
    \item \textbf{Dynamically Generated Hate Speech Dataset}~\cite{vidgen2020learning}: The dataset comprises 41,255 entries generated by a human-and-model-in-the-loop process. It captures hate speech against the ten most frequently targeted groups, including black people, women, and Jews, among others.
    \item \textbf{ToxiGen}~\cite{hartvigsen2022toxigen}: Created using GPT-3, this dataset contains 274,186 machine-generated statements, with over 135k toxic and 135k benign statements. It targets 13 minorities, such as Blacks, Jews, and LGBTQ+ people. 
\end{itemize}

\subsection{Implementation Details}
We used a pre-trained language model (RoBERTa-base~\cite{liu2019roberta} and BERT-base-uncased~\cite{kenton2018bert}) as a sentence encoder in all experiments and used the Adam optimizer with a learning rate of 2e-5, batch size of 16, and dropout of 0.1 during the fine-tuning process. 
We used NVIDIA RTX 4090 GPU (24GB) for training all models, and the hyper-parameters were set to loss scaling parameter $\lambda$ $\in$ \{0.1\}, momentum parameter $m$ $\in$ \{0.999\} based on previous studies, and the temperature parameter $\tau$ $\in$ \{0.05, 0.07, 0.1\}, momentum queue size $q$ $\in$ \{512, 1024, 2048\}, and hard negative $k$ $\in$ \{16, 32, 64\} for further hyper-parameter search. We chose the best model score with macro F1-score in the validation. 


\subsection{Baseline Models}
\begin{itemize}[leftmargin=0.35cm]
\setlength{\itemsep}{0pt}
    \item \textbf{Cross-Entropy (CE) loss}: CE loss is widely adopted as a general approach to classification tasks and hate speech detection.
    \item \textbf{Supervised Contrastive Learning (SCL) with CE loss}~\cite{gunel2020supervised}: This method enhances CE loss by combining supervised contrastive learning. This is effective for complex tasks such as hate speech detection, where distinguishing subtle class differences.
    \item \textbf{Contrastive Learning using Implication (ImpCon) with CE loss}~\cite{kim2022generalizable}: This method improves CE loss by integrating implication-based contrastive learning. ImpCon improves the model to understand contextual relationships by injecting common implications of implicit hate speech.
\end{itemize}

\section{Results and Analysis}\label{results}
\subsection{Implicit Hate Speech Detection Results}
Table~\ref{table_1} shows the evaluation results in the four datasets for the training model on the IHC and SBIC datasets. External knowledge means that additional training knowledge beyond the existing data is used, such as implication and synonym substitution. In the case without external knowledge, the encoder dropout noise~\cite{gao2021simcse} is used as the positive sample. In the cases with external knowledge (Ours, ImpCon), implication was used as the positive sample for hate speech, and synonym substitution was used as the positive sample for non-hate speech, the same as~\citet{kim2022generalizable}.

\begin{figure*}
    \centering
    \includegraphics[width=2.0\columnwidth]{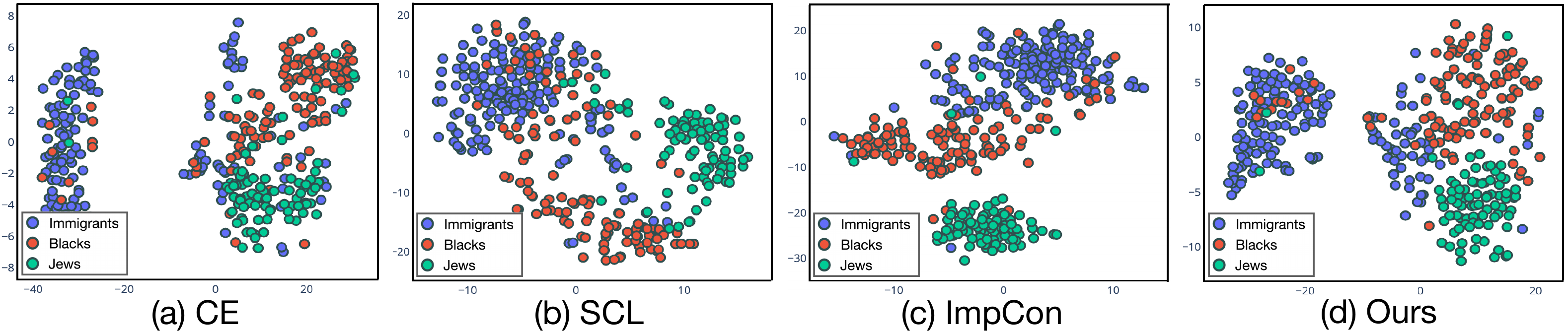}
    \caption{Visualization of implicit hate speech on three targets with the IHC dataset in the In-dataset setting. (Blue: hate speech targeted Immigrants, Red: hate speech targeted Blacks, Green: hate speech targeted Jews.)}
    ~\label{fig4}
    \vspace{-0.45cm}
\end{figure*}
\begin{figure*}
    \centering
    \includegraphics[width=2.0\columnwidth]{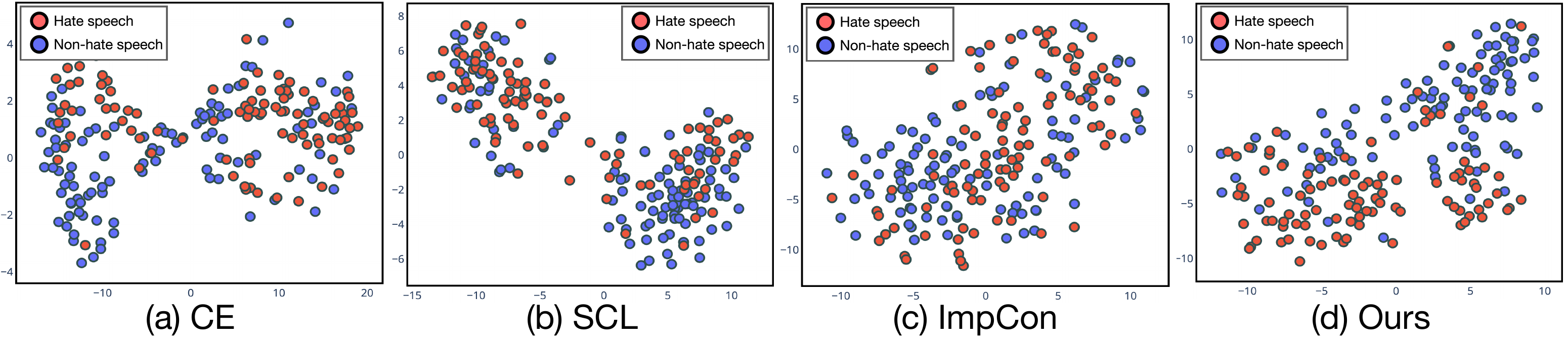}
    \caption{Visualization of implicit hate speech and non-hate sentences that targeted Blacks with the ToxiGen dataset in the zero-shot setting. (Blue: Non-hate speech, Red: Implicit hate speech.)}
    ~\label{fig5}
    \vspace{-0.4cm}
\end{figure*}

In-dataset evaluation (i.e., IHC training set $\rightarrow$ IHC test set, SBIC training set $\rightarrow$ SBIC test set) and cross-dataset evaluation (i.e., IHC training sett $\rightarrow$ all test sets except IHC test set, SBIC training sett $\rightarrow$ all test sets except SBIC test set), LAHN achieves the best performance in 29 out of 32 comparison cases. Furthermore, unlike previous studies~\cite{kim2022generalizable} that found inferior or equal performance to single CE loss learning in the in-dataset case, LAHN achieves better performance than CE loss learning in all in-dataset evaluations. 

LAHN is also robust in the absence of external knowledge. We observe that SCL loss without external knowledge performs poorly relative to CE loss, and ~\citet{kim2022generalizable} shows the same trend for SCL with augmentation based on synonym substitution. In contrast, LAHN, without external knowledge, which uses only dropout noise applied to anchors as a positive sample, achieves the best performance in 15 out of 16 cases without external knowledge comparison. 

In the case using the SBIC training dataset, the average score of LAHN without external knowledge (Average column) outperforms the average score of LAHN with external knowledge in both BERT and RoBERTa models. This indicates that the implication used as external knowledge is of poor quality or that the implication may lead to negative bias in model training. The average scores for the cases without external knowledge using BERT and RoBERTa are 69.75 and 71.28, respectively, while the cases with external knowledge are lower (68.85 and 70.76). This shows that the implication information used as external knowledge for SBIC's implicit hate speech data or the synonym substitution augmentation used as external knowledge for non-hate speech is likely to be of poor quality. 

\subsection{Qualitative Analysis}
Figure~\ref{fig4} shows a visualization of some implicit hate speech data from the IHC validation set using t-SNE~\cite{van2008visualizing} on a model trained on the IHC dataset by three targets (Immigrants, Blacks, Jews). 

SCL (Figure~\ref{fig4}-b) shows more dense clusters compared to CE (a) but still has some samples mixed in with each other. ImpCon and LAHN, on the other hand, form clearer cluster boundaries. We note that LAHN forms different sub-clusters within the immigrant cluster. SCL and ImpCon were trained to cluster similar classes using label information and implication, respectively. They push or pull semantically similar or dissimilar samples within randomly sampled mini-batches, depending on the training strategy. In this process, contrastive loss can degrade the semantic similarity of a representation by further attracting positive samples that are already close enough or pushing negative samples that are already far enough away.

In contrast, our method assumes that the pre-trained language model already has a high-quality representation and encourages the model to use only hard negatives for training in order to increase the classification performance of the model without compromising the semantic information as much as possible. As shown in Figure~\ref{fig4}-d, LAHN ensures the margin between sentences with different target information while maintaining the margin of data with different semantic characteristics within the immigrants.

Figure~\ref{fig5} shows a visualization of the extracted embeddings for the part of the ToxiGen dataset in the zero-shot setting with the model fine-tuned on the IHC dataset. We randomly sampled both hate speech and non-hate speech about black people. The embeddings using LAHN form a relatively sharper boundary between hate speech and non-hate speech in compared to the other three models. This result shows that our LAHN leads the model to form a more generalizable representation compared to the other methods.

\subsection{Effect of Hyperparameters}
\begin{figure}
    \centering
    \includegraphics[width=1.0\columnwidth]{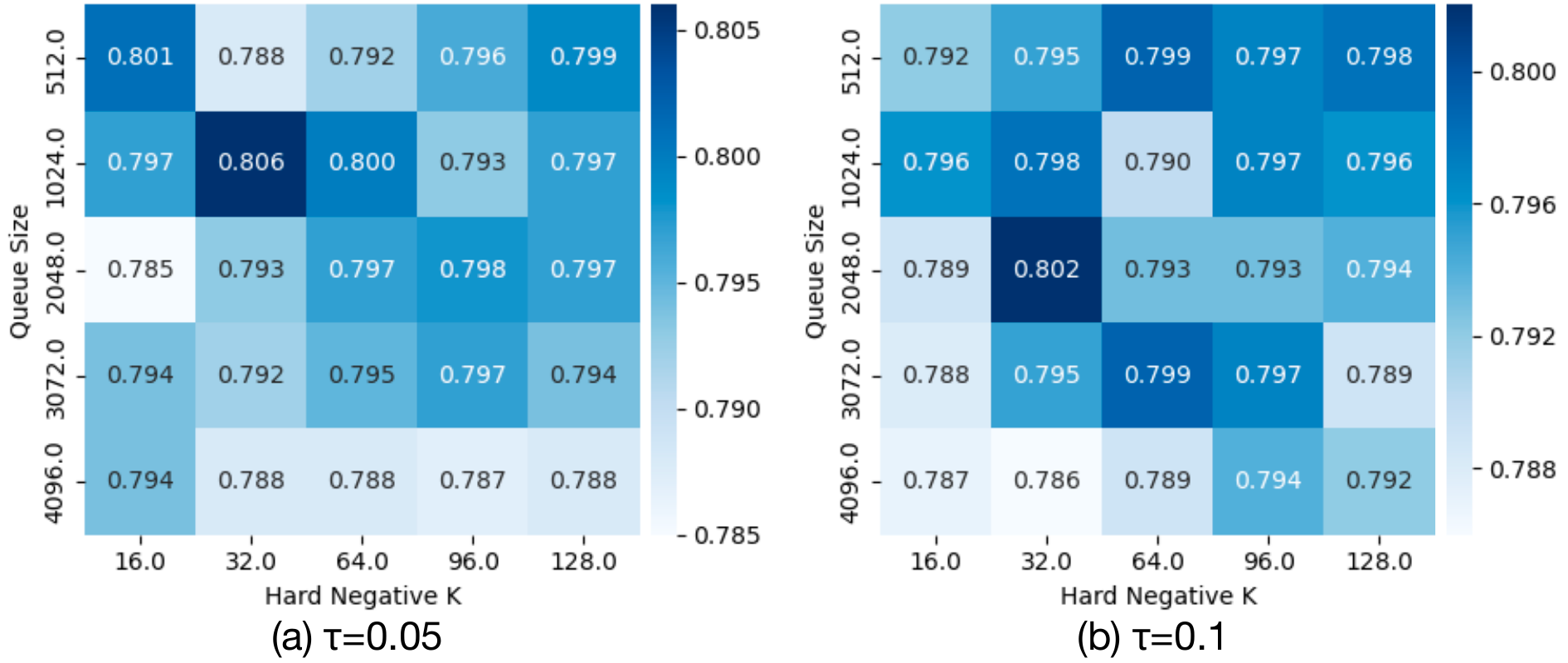}
    \caption{The impact of hyper-parameters $q$ (queue size) and $k$ (sampling size) on IHC dataset with RoBERTa-base model.}
    ~\label{fig6}
    \vspace{-20pt}
\end{figure}

We evaluate the effects of the hyper-parameters $q$ (queue size) and $k$ (hard negative sample size). ~\citet{cao2022exploring} found that for NLP tasks where the model has a fast update rate, a queue size as high as 4096 can cause performance degradation. This shows that, unlike MoCo, queue expansion does not inevitably lead to better performance.
This finding is consistent with our results, and Figure~\ref{fig6} shows the variation of the model's performance with queue size and hard negative sample size. The $x$-axis is the number of hard negatives sampled for each anchor, and the $y$-axis is the size of features that can be candidates for hard negatives.

In contrast to previous work~\cite{cao2022exploring,wu2022esimcse}, we did not use all of the queue features as negative samples but only sampled some of them, meaning that an increase in queue size does not imply the use of outdated information. Therefore, an increase in queue size does not necessarily lead to a linear decrease in performance. Nevertheless, our ablation results show that performance degradation occurs at a certain point in the queue size. That is, as the queue size increases, outdated samples may also become candidates for hard negative sampling, which can negatively affect performance. This result shows that maintaining a large number of hard negative sample candidates does not always improve performance.

Similarly, we found that increasing the number of hard negative samples does not result in a linear performance improvement. Since our method determines the hard negative samples based on the sorted similarity between the anchor and the negative sample, selecting a large number of hard negative samples implies that a high similarity to the anchor cannot be guaranteed. Therefore, our results show that it is necessary to experimentally select the appropriate number of hard negative samples and queue size.

\subsection{Ablation Study} \label{sec:ablation}
\begin{table}[]
\centering
\caption{Ablation study results for in-dataset evaluation of our methods (w/ RoBERTa-base) on IHC and SBIC.}
\label{ablation}
\setlength\tabcolsep{2.0pt}
\resizebox{0.5\textwidth}{!}{
\begin{tabular}{ccc||cc|cc|c}
\hline
\multicolumn{3}{c||}{\textbf{Components}} & \multicolumn{2}{c|}{\textbf{IHC}} & \multicolumn{2}{c|}{\textbf{SBIC}} & \multirow{2}{*}{\textbf{Average}} \\
\cline{1-7}
\textbf{MoCo} & \textbf{HN-Samp} & \textbf{S-Weight} & \textbf{Acc} & \textbf{F1} & \textbf{Acc} & \textbf{F1} &  \\ \hline
\redxmark & \redxmark & \redxmark & 83.12 & 79.08 &   85.72 & 85.39 & 83.33 \\ 
\bluecheck & \redxmark & \redxmark & 83.47 & 78.99 &   84.93 & 84.69 & 83.02 \\ \hline \hline
\bluecheck & \bluecheck & \redxmark & 83.58 & 79.95 &  85.40 & 84.99 & 83.48 \\ \hline \hline
\bluecheck & \bluecheck & \bluecheck & \textbf{84.36} & \textbf{80.58} &  \textbf{86.06} & \textbf{85.80} & \textbf{84.20} \\ \hline
\end{tabular}
}
\end{table}

We conducted the ablation studies by removing each component of our methods on the IHC dataset and the SBIC dataset with RoBERTa-based model. 
\begin{itemize}[leftmargin=0.35cm]
\setlength{\itemsep}{-1pt}
    \item \textbf{MoCo}: \underline{Momentum Contrast} means that all samples in the Momentum Contrast queue with label information are used during contrastive learning. 
    \item \textbf{HN-Samp}: \underline{Hard Negative Sampling} means to use the Label-aware Hard Negative Sampling strategy with the momentum queue. 
    \item \textbf{S-Weight}: \underline{Similarity Weight} means using the prediction probability from the momentum encoder to induce the model to select features with a high probability of confusion as hard negatives.
\end{itemize}

The results are shown in Table~\ref{ablation}. Because of the interdependence of our methods, a total of four progressive ablation studies were performed (i.e., HN-Samp cannot be used alone without SupMoCo, and S-Weight cannot be used alone without HN-Samp). These results show that LAHN can significantly increase all performance compared to the MoCo, including false negatives to increase the number of negative samples, or HN-Samp which only similarity-based hard negative sampling without S-weight.

\section{Conclusion and Future Work}
In this paper, we propose LAHN that incorporates the momentum queue to extract hard negatives, which are more likely to be confused by the model as compared to anchors. We demonstrated the effectiveness of LAHN in both in-dataset and cross-dataset performance evaluations, compared to existing methods.
In the future, we will focus on finding new sampling metrics that are more advanced than similarity-based hard negative sampling for implicit hate speech detection and finding ways to exploit better semantic features from hard negatives.

\section{Limitations}
While our proposed LAHN has demonstrated its effectiveness in implicit hate speech detection, our research has some limitations. 

First, LAHN relies on supervision, which reduces the advantage of MoCo in allowing a large amount of unlabeled data to be used for supervised learning. In addition, the dual encoder framework, which uses an additional MoCo encoder, still has cost limitations compared to other deep learning methods. 

Second, our strategy is limited in that it requires the exploration of a large number of hyperparameters. As we have seen in Section~\ref{sec:ablation}, LAHN exhibits dynamic performance depending on hyperparameters such as $q$, $k$, and $\tau$, which force the cost of hyperparameter exploration to vary across the dataset. 

Third, since we sampled only a fraction of the MoCo queue, we may lose the benefit of MoCo's use of large, consistent negative samples for training. In addition, since we are not using a large queue size like MoCo, it may be more advantageous to adopt an in-batch hard negative sampling strategy. However, we emphasize that our methodology overcomes the constraints of limited computational resources, which may inspire future work.

In future work, we can explore methods that are robust to multiple hyper-parameters, and investigate strategies for effectively sampling out-of-batch hard negatives for training without introducing MoCo.

\section{Ethics Statement}
Our work aims to extend previous research on implicit hate speech detection and contribute to solving social conflicts caused by hateful content. We use publicly available open datasets containing implicit hate speech without compromising user privacy.

Implicit hate speech detection models can still be biased, and our method may cause the risk of increasing false predictions that have not been observed in previous studies and are not revealed by the performance. However, we believe that our ongoing efforts to improve implicit hate speech detection can help mitigate this risk.

\section*{Acknowledgements}
This research was supported by Institute of Information \&
communications Technology Planning \& Evaluation (IITP) grant funded by
the Korea government(MSIT) (No.RS-2020-II201373, Artificial
Intelligence Graduate School Program(Hanyang University)), and the National Research Foundation of Korea (NRF) (2018R1A5A7059549, 2021S1A5A2A03065899). 

\bibliography{anthology,custom}
\bibliographystyle{acl_natbib}

\appendix
\section{Implicit Hate Speech Detection on Large Language Models}\label{sec:appendix}
\subsection{Implementation Details}

To evaluate the performance of Large Language Models on implicit hate speech detection, we employ four LLMs (GPT-3.5-turbo-0125 from OpenAI, Claude-3-Haiku-20240307 from Anthropic, and Llama3-8B-Instruct-v1 and Llama3-70B-Instruct-v1 from Meta). We used the OpenAI API~\footnote{~\href{https://openai.com/api/}{https://openai.com/api/}} for the GPT-3.5-turbo-0125, and used Bedrock on Amazon Web Services~\footnote{~\href{https://aws.amazon.com/bedrock/}{https://aws.amazon.com/bedrock/}} for the other models. 

\subsection{Experimental Results}


\begin{table}[]
\centering
\caption{The performance of four Large Language Models using Zero-Shot (ZS) and Few-Shot (FS) prompts on two evaluation datasets (IHC, SBIC).}
\label{LLMs}
\resizebox{0.475\textwidth}{!}{
\begin{tabular}{c|c|cc}
\hline
\multirow{2}{*}{\textbf{Model}} & \multirow{2}{*}{\textbf{Method}} & \multicolumn{2}{c}{\textbf{Evaluation Dataset}} \\ \cline{3-4} 
                                         &    & \multicolumn{1}{c|}{\hspace*{2.95mm}\textbf{IHC}\hspace*{2.95mm}} & \textbf{SBIC} \\ \hline \hline
\multirow{2}{*}{GPT-3.5-turbo-0125}      & ZS & \multicolumn{1}{c|}{72.02}        & 73.99         \\ \cline{2-4} 
                                         & FS & \multicolumn{1}{c|}{75.30}        & 76.10         \\ \hline
\multirow{2}{*}{Claude-3-Haiku-20240307} & ZS & \multicolumn{1}{c|}{15.83}        & 26.71         \\ \cline{2-4} 
                                         & FS & \multicolumn{1}{c|}{72.20}        & 67.77         \\ \hline
\multirow{2}{*}{Llama3-8B-Instruct-v1}   & ZS & \multicolumn{1}{c|}{37.35}        & 37.62         \\ \cline{2-4} 
                                         & FS & \multicolumn{1}{c|}{73.22}        & 72.36         \\ \hline
\multirow{2}{*}{Llama3-70B-Instruct-v1}  & ZS & \multicolumn{1}{c|}{73.86}        & 57.82         \\ \cline{2-4} 
                                         & FS & \multicolumn{1}{c|}{76.98}        & 73.15         \\ \hline
\end{tabular}
}
\end{table}
Table~\ref{LLMs} shows the performance of the LLMs for the implicit hate speech detection on two datasets (IHC, SBIC). Compared to traditional language models, LLMs with much larger parameters require enormous resources for fine-tuning. Therefore, in-context learning is essential for the efficient use of LLMs~\cite{dong2022survey}. Zero-shot learning, which provides only a description of the task, and Few-shot learning, which provides examples of the task, are the typical in-context learning methods available for LLMs~\cite{brown2020language}.

To ensure a fair few-shot setup, we randomly extracted one hate and one non-hate sample from the original training dataset for each evaluation dataset and fed them as the few-shot samples. We used a randomly sampled subset as test data, similar to previous studies~\cite{roy2023probing, zhang2024don}, to maintain the same number of positive and negative samples and to overcome the cost of LLMs. Finally, we used an IHC and SBIC test dataset of 1K with equal sample proportions.

In all experiments with the four LLMs, few-shot learning consistently outperforms zero-shot learning. On Llama3-8B and Claude3-Haiku, which are known to be relatively lightweight models, zero-shot exhibited F1-scores ranging between 15.83 and 37.62, but after a few-shot sample was injected, performance increased to between 67.77 and 73.22. This suggests that the few-shot prompting strategy is effective for implicit hate speech detection.

From a cost perspective, LLMs still require a large amount of resources and have lower performance compared to fine-tuned language models that have been trained on the specific dataset. Therefore, we argue that our method of training specialized detection models at a relatively low cost still has advantages over in-context learning with LLMs.

\end{document}